\pdfoutput=1

\documentclass[11pt]{article}

\usepackage{ACL2023}

\usepackage{times}
\usepackage{latexsym}

\usepackage[T1]{fontenc}

\usepackage[utf8]{inputenc}

\usepackage{microtype}

\usepackage{inconsolata}

\usepackage{multirow}
\usepackage{makecell}

\usepackage{times}
\usepackage{latexsym}
\usepackage{graphicx}
\usepackage{subfigure}
\usepackage{float}
\graphicspath{{images}}
\usepackage{stfloats}
\usepackage{booktabs}
\usepackage{indentfirst}
\usepackage{amsmath}
\usepackage{amssymb}

\usepackage{enumitem}

\usepackage{colortbl}
\usepackage{arydshln}

\usepackage[linesnumbered,ruled,vlined]{algorithm2e}

\SetCommentSty{mycommfont}

\SetKwInput{KwInput}{Input}                
\SetKwInput{KwOutput}{Output}              

\usepackage{CJKutf8}

\usepackage{listings}
\usepackage{xcolor}

\definecolor{codegreen}{rgb}{0,0.6,0}
\definecolor{codegray}{rgb}{0.5,0.5,0.5}
\definecolor{codepurple}{rgb}{0.58,0,0.82}
\definecolor{backcolour}{rgb}{0.95,0.95,0.92}

\lstdefinestyle{mystyle}{
    backgroundcolor=\color{backcolour},   
    commentstyle=\color{codegreen},
    keywordstyle=\color{magenta},
    numberstyle=\tiny\color{codegray},
    stringstyle=\color{codepurple},
    basicstyle=\ttfamily\small,
    breakatwhitespace=false,         
    breaklines=true,                 
    captionpos=b,                    
    keepspaces=true,                 
    numbers=left,                    
    numbersep=5pt,                  
    showspaces=false,                
    showstringspaces=false,
    showtabs=false,                  
    tabsize=2
}

\title{Pursuing Best Industrial Practices for Retrieval-Augmented Generation in the Medical Domain}

\author{
    Liz Li\textsuperscript{1,}, Wei Zhu\textsuperscript{2,}\thanks{\ \ Corresponding author. For any inquiries, please contact: michaelwzhu91@gmail.com. }\\
    \small \textsuperscript{1}\ DataSelect AI, Xuhui, Shanghai, China \\
    \small \textsuperscript{2}University of Hong Kong, Hong Kong, HK, China
}

\begin{document}
\maketitle
\begin{abstract}

While retrieval augmented generation (RAG) has been swiftly adopted in industrial applications based on large language models (LLMs), there is no consensus on what are the best practices for building a RAG system in terms of what are the components, how to organize these components and how to implement each component for the industrial applications, especially in the medical domain. In this work, we first carefully analyze each component of the RAG system and propose practical alternatives for each component. Then, we conduct systematic evaluations on three types of tasks, revealing the best practices for improving the RAG system and how LLM-based RAG systems make trade-offs between performance and efficiency. 

\end{abstract}

\begin{CJK*}{UTF8}{gbsn}

\section{Introduction}

Large Language Models (LLMs) have transformed the way people access information online, shifting from conventional web searches to direct interactions with chatbots. Recently, there has been a rapid development of commercial LLMs by industry players, demonstrating state-of-the-art performance in question answering (QA) across both general and medical domains \cite{achiam2023gpt4,anil2023palm2,singhal2023large,singhal2023towards,nori2023capabilities,Cui2023UltraFeedbackBL,wang2024ts,yue2023-TCMEB,Zhang2023LearnedAA,2023arXiv230318223Z,Xu2023ParameterEfficientFM,Ding2022DeltaTA,Xin2024ParameterEfficientFF,qin2023chatgpt,PromptCBLUE,text2dt_shared_task,Text2dt,zhu_etal_2021_paht,Li2023UnifiedDR,Zhu2023BADGESU,Zhang2023LECOIE,Zhu2023OverviewOT,guo-etal-2021-global,zhu-etal-2021-discovering,Zheng2023CandidateSF,info:doi/10.2196/17653,Zhang2023NAGNERAU,Zhang2023FastNERSU,Wang2023MultitaskEL,Zhu2019TheDS,zhu2021leebert,Zhang2021AutomaticSN,Wang2020MiningIH,li2025ft,leong2025amas,zhang2025time,yin2024machine,zhu2026mrag,zhu2026evaluatechatgpt}. Despite these achievements, a notable issue is that LLMs can sometimes produce responses that, while plausible, are factually inaccurate, a problem referred to as hallucination \cite{ji2023survey,rawte2023survey}. Additionally, the training data for these models may not include the most up-to-date information, such as recent news or the latest scientific and medical research. These shortcomings present significant challenges and potential risks, particularly in critical areas like finance, personal assistance, bio-medicine and healthcare \cite{tian2024opportunities,hersh2024search,tian2024opportunities,hersh2024search,zhu2024iapt,zhu-tan-2023-spt,Liu2022FewShotPF,xie2024pedro,Cui2023UltraFeedbackBL,zheng2024nat4at,zhu2023acf,gao2023f,zuo-etal-2022-continually,zhang-etal-2022-pcee,sun-etal-2022-simple,zhu-etal-2021-gaml,Zhu2021MVPBERTMP,li-etal-2019-pingan,zhu2019panlp,zhu2019dr,zhou2019analysis,zhang2025time,wang2025ts,liu2025parameter,yi2024drum,tian2024fanlora}.

\begin{figure*}[t]
\centering
\includegraphics[width=0.66\textwidth]{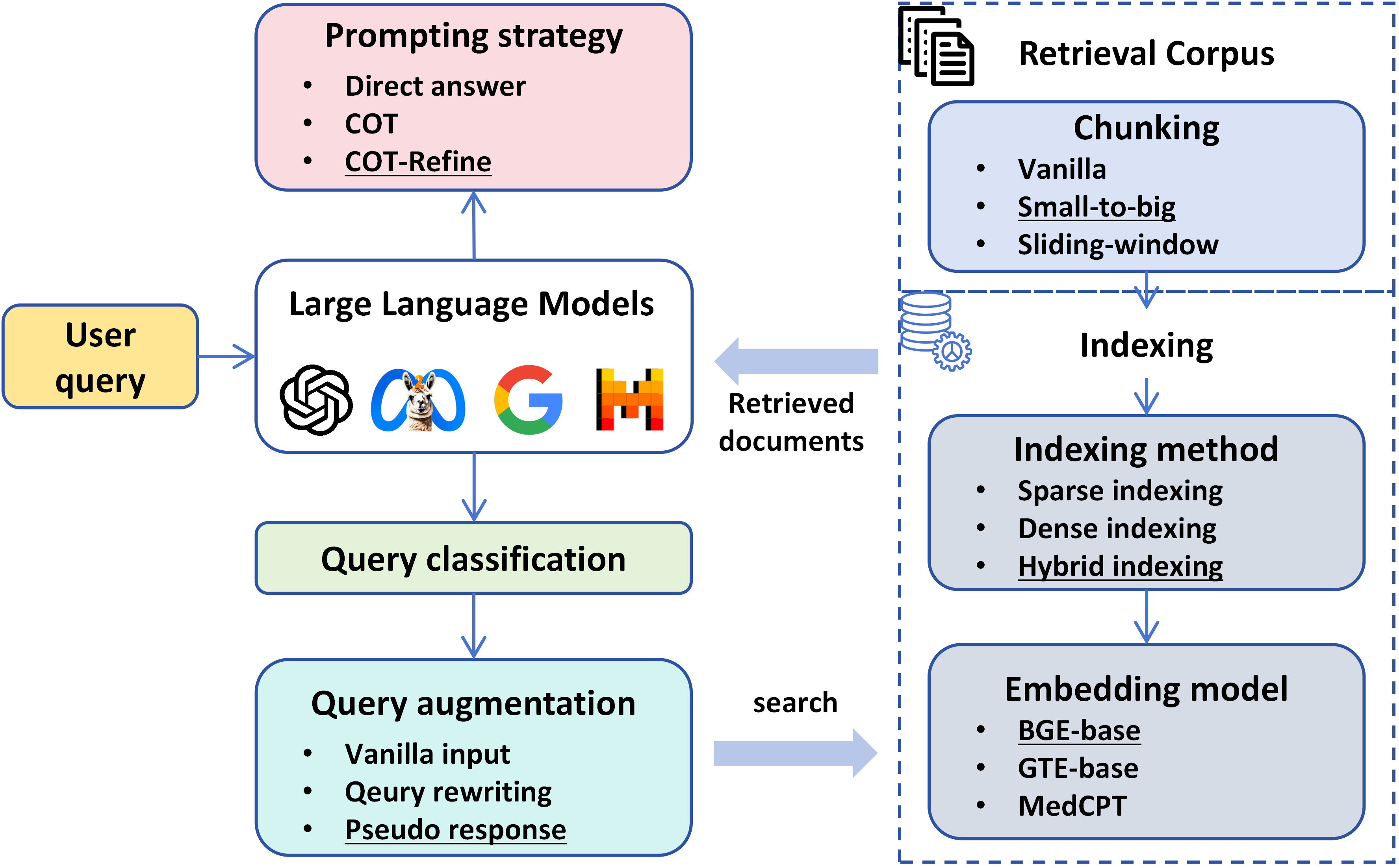} 
\caption{The workflow of retrieval-augmented generation considered in this study. We investigate the contribution of each module and provide insights into optimal RAG practices through extensive experiments. The methods selected for the best practice are \underline{underlined}.}
\label{fig:architecture}
\end{figure*}

To tackle this issue, retrieval-augmented generation (RAG) utilizes current and trustworthy document collections to boost the capabilities of Large Language Models (LLMs), potentially overcoming various challenges in the field \cite{lewis2020retrieval,gao2023retrieval,zhao2024retrieval}. By anchoring LLMs' reasoning in these retrieved documents, RAG can also improve their explainability and transparency. As illustrated in Figure \ref{fig:architecture}, a standard RAG system for the open-domain or medical domain question answering typically involves several key steps: (a) query classification, which assesses whether a given input requires retrieval; (b) construction of the retrieval corpus, encompassing processes like chunking and indexing; (c) the retrieval process, which identifies the most relevant information based on the search input; and (d) response generation, where prompting strategies are used to guide the LLMs. The complexity and challenge lie in the variability of approaches for each step. For instance, when retrieving pertinent documents for an input, multiple techniques such as query rewriting \cite{ma2023query} and pseudo-response generation \cite{gao2022precise} can be applied to enhance the original query for more effective searching. Thus, one central research question raises:

\noindent\emph{\textbf{RQ1.} What is the best practice for building a RAG system, especially for the bio-medical tasks? }

This study is designed to search for the best practices of RAG systems through comprehensive experimentation on both open-domain and medical domain tasks. Given the impracticality of evaluating every possible combination of methods, we employ a three-step strategy to pinpoint the most effective RAG practices. Initially, we examine representative methods for each step or module within the RAG process. Then, we optimize one module at a time while keeping the others constant, iterating through all modules to establish an optimal configuration for the RAG system. Lastly, we showcase the experimental outcomes of this optimal RAG setup and also present variations by altering one module at a time, thus generating a series of RAG configurations. Based on these results, we propose several strategies for deploying RAG that effectively balance performance and efficiency.

In summary, our contributions are two-fold:
\begin{itemize}
\item We thoroughly investigate existing approaches for different modules of the RAG system. 

\item We have conducted extensive experiments to investigate many combinations for the RAG settings and identify the optimal RAG practices. 

\end{itemize}

\section{Related work}

\subsection{Retrieval-augmented generation}

Retrieval-Augmented Generation (RAG) was proposed by \citet{lewis2020retrieval} to enhance the generation performance on knowledge-intensive tasks by integrating retrieved relevant information. In the LLM era led by OpenAI's ChatGPT and GPT-4, RAG not only mitigates the problem of hallucinations as LLMs are grounded on given contexts but can also provide up-to-date knowledge that the LLMs might not encode \cite{gao2023retrieval,zhao2024retrieval}. Many recent studies have been devoted to improving upon the vanilla RAG workflow by either designing novel retrieval and generation mechanisms \cite{borgeaud2022improving,zhang2023repocoder,ram2023context,jiang2023active}, or incorporating pre-training and fine-tuning for improving LLMs' capabilities in RAG \cite{zhang2024raft,siriwardhana2023improving,xue2024badrag}.

In specific domains like bio-medicine, current systematic evaluations of LLMs typically focus on the vanilla LLMs without RAG \cite{PromptCBLUE,singhal2023large,singhal2023towards,nori2023capabilities,chen2023large,saab2024capabilities}. There has been a series of works on how RAG can help to improve LLMs' capabilities in tasks like clinical decision-making, scientific literature analysis, and information extraction \cite{frisoni2022bioreader,naik2021literature,xiong2024benchmarking,lala2023paperqa,jin2024agentmd,zakka2024almanac,jeong2024improving,wang2023augmenting}. However, (a) comprehensive evaluation that contains a variety of tasks is lacking, and (b) systematic investigations on what are the best practices to build a RAG system in these domains, such as the prompt strategies, are lacking, hindering its further industrial applications. Our work complements the existing literature by investigating best practices for the LLM-based RAG system.

\section{RAG system }
\label{sec:rag_system}

In this section, we detail the components of the RAG system, and the approaches available. Figure \ref{fig:architecture} presents the RAG system with different methods for each component.

\subsection{Query classification}

In an RAG system, not all queries necessitate retrieval augmentation and can be effectively handled by the inherent capabilities of LLMs. Although RAG improves information accuracy and mitigates hallucinations, it also introduces increased latency and computational complexity. Thus, our approach begins with classifying queries to ascertain whether retrieval is needed. Queries that require additional information go through the RAG modules, while those that do not are directly processed and answered by the LLMs.

Therefore, we propose the classifying task of determining if a query needs retrieval, and we train a classifier to automate this decision-making process. To build the dataset for this task, we label the query in the dataset as \emph{need RAG} if its answer's log-likelihood function can increase when conditioned on retrieved documents. Otherwise, the query will be labeled \emph{not need RAG}.

\subsection{Chunking}

It is essential to properly chunk the documents into smaller segments to enhance retrieval precision and avoid length issues in LLMs. In this study, we use the sentence-level chunking strategy \cite{gao2023retrieval}, which is to break documents into sentences and fill the sentences into chunks. During chunking, an important aspect to consider is the chunk size. Chunk size significantly impacts performance. Larger chunks provide more context but could introduce irrelevant information and increase latency. Smaller chunks improve retrieval recall and efficiency but may lack sufficient context. This study considers the chunk size $L_{c} = 256$ by default. Moreover, we must consider the following aspect of chunking, which determines how neighboring chunks are connected.

\noindent \textbf{Chunking techniques} \quad We consider the following three chunking techniques: (a) Vanilla chunking, which chunks the documents into non-overlapping segments of length $L_{c}$. (b) Small2big chunking, which uses a small segment from vanilla chunking for retrieval, but a longer segment containing the small one will be returned. We set the length of the smaller chunk to $L_{c}^{s} = L_{c} / 2$, and the larger chunk to $L_{c}^{l} = L_{c} $. (c) Sliding-window chunking, which chunks the documents into overlapping segments. The overlap length is set to $L_{o} = L_{c} / 4$.

Intuitively, the latter two techniques improve retrieval quality over the vanilla chunking method by organizing chunks' relationships.

\subsection{Document indexing}
\label{subsec:indexing}

After long documents are chunked into smaller segments, we save them in a database with an index built for efficient retrieval. Based on how the documents are indexed, the document indexing methods are: (a) Sparse indexing, which utilizes the technique of inverted index. It first tokenizes the document segments into words or tokens and then builds an inverted index where each key term points to a list of documents that contain that term. (b) Dense indexing, which assumes the document segments are transformed to semantic vectors via an embedding model. Approximate nearest neighbor (ANN) index \cite{abbasifard2014survey} like HNSW or IVF is the possible method for indexing such a vector database. (c) Hybrid indexing, which is to build both indexes on the corpus.

\noindent \textbf{Embedding models} \quad Since we consider dense indexing, a high-quality embedding model is central to the retrieval performance. In this work, we consider a series of embedding models, both open-domain and in-domain, with base sizes: (a) BGE-base \cite{xiao2023c}. (b) GTE-base \cite{li2023towards}. (c) MedCPT \cite{jin2023medcpt}, a medical document representation model initialized from PubMedBERT \cite{gu2021domain}, and further pretrained on the retrieval tasks curated with PubMed.

\subsection{Retrieval}
\label{subsec:retrievers}

Given a user's query, the retrieval module selects the top $k$ most relevant documents from a pre-built corpus, based on the similarity between the query and the documents. The generation model subsequently uses these selected documents to formulate an appropriate response to the query. How the corpus is indexed directly determines the retrieval algorithm: (a) Best Matching 25 (BM25) \cite{robertson2009probabilistic} algorithm is used for sparse indexing. (b) ANN search is used for dense retrieval. (c) hybrid search is used if the corpus uses hybrid indexing. Sophistic toolkits are available to implement the first two. To implement a hybrid search, one first uses the former two methods and then combines the search results with a 3:1 weight ratio for similarity scores to rerank the two lists of retrieved documents.  

\noindent \textbf{Query augmentation strategy for search} \quad In a standard RAG system, the user query is input into the search engine to retrieve relevant documents. However, the original queries often have poor expression and lack semantic information \cite{gao2022precise}, negatively impacting the retrieval results. Thus, we evaluate four approaches to search inputs: (a) Vanilla input, that is, directly utilizing the user query as search input. (b) Query rewriting, that is, refining queries and transforming the queries to sub-questions so that they can better match relevant documents. (c) Pseudo-response generation, that is, the model first generates a response, which will be concatenated to the user query.

\subsection{Response generation}
\label{subsec:prompting_strategies}

\noindent \textbf{prompting strategy} \quad When the retrieval procedure returns a list of referential documents, an LLM will concatenate these contents at the front of the prompt and generate the final response. Other than relevant documents, the prompt contains instructions that reflect the prompting strategies: (a) Direct answer (DA): Given the question, the prompt asks the LLM to output the answer directly. (b) Chain-of-thought (COT) \cite{Wei2022ChainOT} explicitly asks the LLM to think step by step and demonstrate the intermediate outputs. (c) COT-Refine. Building on COT and Self-Refine\cite{madaan2024self}, this strategy assumes a response without RAG has been generated, and this prompt asks the model to reflect on this response, utilize the retrieved documents to make necessary corrections, and finally draft a new response.

\section{Experiments}
\label{sec:experiment_results}

We systematically evaluate the RAG workflow on a series of knowledge-intensive tasks, providing a multidimensional analysis of different components in RAG and paving the way for optimal practices for implementing RAG.

\subsection{Experimental settings }

\noindent \textbf{Evaluation datasets} \quad We evaluate the RAG systems on three benchmarks, investigating how RAG helps the LLMs in open-domain or in-domain question-answering tasks and information extraction tasks: (a) MMLU \cite{hendrycks2020measuring}. (b) PubMedQA \cite{jin2019pubmedqa}. (c) PromptNER, a mixture of samples from multiple named entity recognition tasks, including CoNLL-03 \cite{sang2003introduction}, OntoNotes 5.0\footnote{https://catalog.ldc.upenn.edu/LDC2013T19} and BioNLP2004 \cite{collier-kim-2004-introduction}. Introductions and dataset statistics are in Appendix \ref{sec:app_datasets}.

Now we elaborate on the steps we take to construct the dataset for query classification:

\noindent (i) Query collection. We collects: (a) 10k samples from the dev set and test set of the original MMLU datasets, with no overlapping with the test set for RAG evaluation. (b) 10k samples from the automatic labeled set of PubMedQA. (c) 7.9k dev set from PromptNER. Thus, we have a collection of  27.9k samples with query-response pairs. 

\noindent (ii) Query labeling. We conduct automatic labeling of the above dataset using the LlaMA-2 13B model and log-likelihood function. For a sample with a query-response pair, we first calculates the log-likelihood of the response text conditioned only on the query, that is, $l_0 = \text{LLL}(\text{response} | \text{query})$. Here, $\text{LLL}()$ is the log-likelihood function calculated with the given LlaMA-2 13B backbone. We search the corpus with hybrid search (as in Section \ref{subsec:retrievers}), and retrieve $k = 8$ documents, denoted as $\text{docs}$. Conditioned on the $\text{docs}$ and $\text{query}$, the response's log-likelihood becomes $l_1 = \text{LLL}(\text{response} | \text{query}, \text{docs})$. The query is labeled "need RAG" (or label 1) if $l_1 - l_0 > 0$, and "not need RAG" (label 0) otherwise. Using the above procedure, there are 17.9\% queries with the "need RAG" label (denoted as label 1). 

\noindent (iii) Dataset split. The automatically labeled queries are split into a 24k:2k:1.9k train/dev/test split.

\noindent \textbf{Evaluation metrics} \quad For the MMLU and PubMedQA tasks, we will directly consider the correctness of the final answers. Thus, we report accuracy (denoted as acc). 

For the PromptNER task, the output response text will first be parsed and transformed to a json instance. If the response can not be parsed to json, then we consider the prediction as a null list. We adopt the instance-level strict micro-F1 following \citet{PromptCBLUE}, that is, the model predicts an entity correctly if and only if it correctly predicts all its components.

Other than the performance matrices on the evaluation datasets, we also measure the efficiency of the RAG systems by the average latency (in seconds (s)) for completing the response for a test sample.

\noindent \textbf{Settings for query classification} \quad We use the query classification dataset introduced above to build a query classifier. The pre-trained backbones we consider are: (a) BERT-base \cite{devlin-etal-2019-bert}, (b) RoBERTa-base \cite{liu2019roberta}, and (c) DeBERTa-base \cite{He2020DeBERTaDB}. For fine-tuning these models, we utilize the HuggingFace Transformers package \cite{Wolf2019HuggingFacesTS}, with batch size 64, learning rate 2e-5, warm-up steps 100, AdamW optimizer \cite{Loshchilov2019DecoupledWD}. The other hyper-parameters are kept the same with Transformers. For every 200 optimization steps, we run evaluations on the dev set and save the model checkpoint. The checkpoint with the best dev accuracy is used as the final fine-tuned model to make predictions on the test set. And this checkpoint is used as a part of the RAG system.

\begin{table}[tb!]
\centering
\resizebox{0.34\textwidth}{!}{
\begin{tabular}{c|cc}
\hline
\textbf{Model backbone }   &     \textbf{Acc}       &    \textbf{F1}  \\ 
\hline
BERT-base  &    0.586   &    0.341    \\
RoBERTa-base   &    0.638   &   0.382   \\ 

DeBERTa-base   &   0.637    &  0.354  \\ 

\hline
\end{tabular}}

\caption{\label{tab:results_query_cls} The performance of query classification.  } 
\vspace{-8pt}
\end{table}

\begin{table*}

\centering
\resizebox{0.84\textwidth}{!}{
\renewcommand\arraystretch{1.2}
\begin{tabular}{cc|ccc|cc}
\hline

Method  &   RAG Setting &  MMLU   &  PubMedQA   &  PromptNER   &  Avg score   &  Avg latency  \\
\hline

\hline

BP-RAG   &   BP-RAG    &    \textbf{59.7}      &  56.9    &   \textbf{25.8}    &     \textbf{47.5}    &     14.3    \\

\hdashline

No RAG   &     -    &   49.3    &    43.4      &  20.6    &    37.8    &     10.8   \\

RAG\_1   &     + vanilla chunking    &    59.3      &   55.9      &    25.2    &  46.7    &    14.1 \\
RAG\_2   &     + sliding-window chunking     &    59.7   &  56.1    &      25.4    &    47.1  &    14.2    \\

\hdashline

RAG\_3   &     + sparse indexing     &     53.1   &    47.3    &   22.5     &         40.9    &    14.2  \\

RAG\_4   &     + dense indexing    &     58.9      &  55.7  &    25.3    &    46.6   &    14.3 \\

RAG\_5   &     + MedCPT     &       55.6     &   \textbf{57.1}  &    24.1      &     45.6   &   14.3  \\
RAG\_6   &     + GTE-base    &      59.3     &  56.2     & 25.7    &    47.1   &    14.2 \\

\hdashline

RAG\_7   &     - query classification   &     58.5  &  55.8    &    25.1     &     46.5    &    20.7      \\

RAG\_8   &     + query rewriting    &      57.4     &  54.5    &    23.8     &   45.2    &    11.4   \\

RAG\_9   &     + vanilla query    &       56.2      &  51.6    &    22.6     &    43.5    &   11.0  \\

\hdashline

RAG\_10   &    + COT    &      58.2     &    55.8     &  24.4    &      46.1    &     14.2    \\
RAG\_11   &    + direct answering    &     54.9    &  51.7 &   21.9    &    42.8      &     3.7   \\

\hline
\end{tabular}}
\caption{ Results of different RAG settings.  The average score is calculated by averaging the scores of all tasks, while the average latency is measured in seconds per query. }
\label{tab:main_results}
\vspace{-12pt}
\end{table*}











\noindent \textbf{Retrieval corpus} \quad The retrieval corpus consists of the following sources: (a) Wikipedia (English) corpus\footnote{\url{https://en.wikipedia.org/wiki/Wikipedia:Database_download}}, a large-scale open-source encyclopedia containing 6.5 million documents for world knowledge. (b) PubMed\footnote{\url{https://pubmed.ncbi.nlm.nih.gov/}} is the most widely used literature resource, containing 23.9 million biomedical articles' titles and abstracts. (c) a proprietary corpus containing 1.3 million books or documents from science, education, and medicine. When processing the retrieval corpus, we set chunk size $L_{c} = 256$.

\noindent \textbf{Retrieval settings} \quad For sparse indexing, we utilize the ElasticSearch toolkit\footnote{https://www.elastic.co/} with the BM25 algorithm for search. For dense indexing, we utilize the Faiss\footnote{https://github.com/facebookresearch/faiss} toolkit to index the document vectors and implement efficient vector search. The top $k = 8$ document segments are retrieved and concatenated to the input prompt (if using RAG for response generation) for each query.

\noindent \textbf{LLM backbone} \quad Our experiments uses the most recent open-sourced LLM, LlaMA-2-chat 13B released by Meta \cite{Touvron2023Llama2O}. After receiving a prompt or instruction, all the predictions are generated using the model's pretrained language modeling head (LM head). For decoding during inference, we use beam search with beam size 3.

\noindent \textbf{Strategy to determine the optimal practice} \quad To begin with, we consider the following settings for the RAG system: sliding-window chunking as the chunking strategy, semantic vector indexing for indexing the retrieval corpus, BGE-base as embedding model, pseudo-response generation for query augmentation, COT for prompting. Following the framework depicted in Figure \ref{fig:architecture}, we optimize individual modules step-by-step, and select the most effective option among the possible choices. This iterative process continued until we could not improve the average task score.

\noindent \textbf{Settings for the RAG system} \quad With the help of the above strategy, we have locked down the optimal practice for the RAG system: small2big chunking as the chunking strategy, hybrid indexing for indexing the retrieval corpus, BGE-base as the embedding model, using query classification, pseudo-response generation for query augmentation, COT-Refine for prompting. We denote this setting as $\text{BP-RAG}$. We also consider the following settings to demonstrate the superiority of $\text{BP-RAG}$: (a) No RAG, which is not to use RAG at all. This setting is presented as a sanity check. (b) RAG\_1, which substitute small2big chunking in $\text{BP-RAG}$ to the vanilla chunking. (c) RAG\_2, which substitute small2big chunking in $\text{BP-RAG}$ to the sliding-window chunking. (d) RAG\_3, which substitute hybrid indexing in $\text{BP-RAG}$ to the sparse indexing. As a result, RAG\_3 will use BM25 for search and not use an embedding model. (e) RAG\_4, which substitute hybrid indexing in $\text{BP-RAG}$ to the dense indexing. (f) RAG\_5, which substitute BGE-base in $\text{BP-RAG}$ to MedCPT, an in-domain embedding model. It is interesting to see whether MedCPT still performs well on open-domain retrieval. (g) RAG\_6, which substitute BGE-base in $\text{BP-RAG}$ to GTE-base. (h) RAG\_7, which does not employ the query classification module. That is, it will retrieve documents for any given query. (i) RAG\_8, which substitute pseudo-response generation in $\text{BP-RAG}$ to query rewriting. (j) RAG\_9, which substitute pseudo-response generation in $\text{BP-RAG}$ to vanilla query. (k) RAG\_10, which substitute COT-Refine in $\text{BP-RAG}$ to COT. (l) RAG\_11, which substitute COT-Refine in $\text{BP-RAG}$ to direct answering.

\subsection{Results}

The results of the RAG settings are presented in Table \ref{tab:main_results}, while the results for the query classification tasks are reported in Table \ref{tab:results_query_cls}. Based on the experimental results presented in Table \ref{tab:main_results}, the following observations can be made:

(a) The benefit of using RAG. Compared with not using RAG and making responses directly, BP-RAG significantly improves the performance by 25.6\% on average. However, one can not ignore that the latency of BP-RAG is 32.4\% higher than No RAG. 

(b) Query classification module. Table \ref{tab:results_query_cls} reports the query classification performance on the query classification dataset we developed. DeBERTa-base \cite{He2020DeBERTaDB} outperforms BERT-base \cite{devlin-etal-2019-bert} and RoBERTa-base \cite{liu2019roberta} by achieving an accuracy of 65.3\%. According to Table \ref{tab:main_results}, this module is beneficial for both the effectiveness and efficiency of the RAG system, leading to an improvement in the average score from 46.5\% to 47.5\% and a reduction in latency time from 20.7 to 14.3 seconds per query. The operations in this module do not significantly increase the overall latency of the system since classifying one query with DeBERTa-base can be done in less than 20 ms.

(c) Chunking strategy. Table \ref{tab:main_results} demonstrates that the small2big strategy slightly outperforms the other two chunking strategies, demonstrating the benefit of retaining contextual information in the retrieved documents.

(d) Indexing Module. The experimental results show that the hybrid indexing strategy attains the highest scores. Hybrid indexing combines the search results from sparse and dense indices, making the retrieved documents more informative.

(e) Embedding models. Among the three base-sized embedding models, the BGE-base model works the best with the RAG. The MedCPT model is further pre-trained on the PubMed corpus, making it especially beneficial for the PubMedQA task, but it is unsuitable for the open-domain MMLU and PromptNER tasks.

(f) Query augmentation module. In this module, query rewriting and pseudo-response generation significantly increase the latency. The former increases the latency by 3.6\%, while the latter increases the latency by 30.1\%. However, pseudo-response generation benefits the RAG system by helping retrieve more relevant and informative documents.

(g) Prompting module. Direct answering is the most efficient in this module, but it significantly underperforms compared to its two competitors. From Table \ref{tab:main_results}, COT-Refine outperforms COT, demonstrating that explicitly asking the LLM to reflect on the past response helps improve its performance.

The experimental results demonstrate that each module contributes uniquely to the overall performance of the RAG system. Note that boosting performance requires the RAG system to increase latency in some of the modules. Thus, in certain time-sensitive applications, industrial practitioners can select settings different from BP-RAG, making an informed trade-off between performance and efficiency.

\section{Conclusion}

This study investigates the best practices for implementing a retrieval-augmented generation (RAG) system. First, we identify potential solutions for each module within the RAG framework. Second, we conduct extensive experiments to systematically assess these solutions and recommend the most effective approach for each module. During this process, we demonstrate how different solutions affect the RAG system's performance and latency. Our findings contribute to a deeper understanding of RAG systems and also provide key insights for industrial applications.

\section*{Limitations}

Despite the fact that we provide extensive experiments for medical RAG on a wide collection of tasks, the following limitations remain: (a) We focus on open-sourced LLMs. Powerful language models like GPT-4o, Gemini \cite{reid2024gemini}, Claude-3\footnote{\url{https://www.anthropic.com/news/claude-3-family}.}, Grok\footnote{https://github.com/xai-org/grok-1} are not evaluated due to resource limitation. (b) There are literature in RAG that adopt more complicated workflow than our RAG system (in Figure \ref{fig:architecture}), such as iterative retrieval \cite{zhang2023repocoder,jiang2023active}. These methods require more LLM inference times. These more advanced RAG strategies have not been evaluated in our current version, but we will address this aspect in our updated version.

\section*{Ethical statement}

Our work's investigations on the best practices of Retrieval-Augmented Generation (RAG) systems presents significant societal benefits alongside critical considerations. By integrating Retrieval-Augmented Generation (RAG) with Large Language Models (LLMs), the RAG framework enhances access to reliable medical information, supporting clinical decision-making and improving patient outcomes through evidence-based responses. The system’s transparency—enabled by source attribution to retrieved documents—helps build trust in AI-assisted medical applications while mitigating "black box" concerns. Our investigations and experimental results provide useful messages to the RAG systems in the industry. 

However, these advancements also present potential risks requiring proactive mitigation. Over-reliance on AI systems could inadvertently erode human clinical judgment, necessitating balanced implementation where RAG serves as an assistive tool rather than a decision-maker. Workforce implications and job displacement risks call for parallel investments in healthcare worker retraining programs. We aim to advance LLM technologies that ethically augment medical expertise while preserving human oversight.

\bibliography{custom}
\bibliographystyle{acl_natbib}

\appendix

\section{Appendix: Datasets}
\label{sec:app_datasets}

The datasets we experiment on are as follows: 
\begin{itemize}
\item \textbf{MMLU} \quad The Massive Multitask Language Understanding (MMLU) benchmark \cite{hendrycks2020measuring} has been introduced to assess the knowledge gained by large language models during pretraining, specifically in zero-shot and few-shot scenarios. This approach aims to make the evaluation process more rigorous and akin to human assessment standards. The MMLU covers 57 diverse topics, spanning STEM, humanities, social sciences, and many others, with questions ranging from elementary school to advanced professional levels. It evaluates not only factual knowledge but also problem-solving skills, covering a wide array of fields from conventional subjects like mathematics and history to more specialized domains such as law and ethics. The extensive range and detailed coverage of these subjects make the MMLU particularly effective for pinpointing areas where a model may struggle. Initially, the dataset comprises a development set of 1,500 samples and a test set of 14.1k samples. For our purposes, we have selected 50 test samples from each of the 57 categories, creating a subset of 2.8k test samples.

\item \textbf{PubMedQA} \quad PubMedQA \cite{jin2019pubmedqa} is a biomedical research QA dataset. The dataset is released under the CC BY 4.0 (Creative Commons Attribution 4.0 International) license and released in \url{https://pubmedqa.github.io/}. It has 1k manually annotated questions constructed from PubMed abstracts. To test the capability of RAG systems to find related documents and answer the question accordingly, we discard the relevant context for each question originally included in the dataset. The possible answer to a PubMedQA question can be yes/no/maybe, reflecting the authenticity of the question statement based on biomedical literature.  This task has a set of 1.0k manually labeled PubMedQA samples, which will be considered the test set for evaluating the RAG system. And the set of 211k automatic labeled samples will be used to construct the dataset for query classification.

\item \textbf{PromptNER} \quad This dataset is a mixture of samples from multiple named entity recognition tasks: 
\begin{itemize}
\item CoNLL-03 \cite{sang2003introduction}. The CoNLL 2003 Named Entity Recognition (NER) dataset, introduced as part of the Conference on Natural Language Learning (CoNLL) shared task in 2003, is a cornerstone benchmark for assessing named entity recognition systems. This dataset features a train/dev/test split of 14,041:3,250:3,453, making it a valuable resource for NLP researchers and practitioners.

\item OntoNotes 5.0\footnote{https://catalog.ldc.upenn.edu/LDC2013T19}. A comprehensive resource in Natural Language Processing (NLP), OntoNotes 5.0 extends its predecessors with rich annotations covering part-of-speech tagging, syntactic parsing, semantic role labeling, and named entity recognition across multiple languages, including English, Chinese, Arabic, and more. For NER, it categorizes entities into predefined classes such as Person, Location, Organization, and Misc (for miscellaneous entities). The dataset's diversity, encompassing various text genres like newswire, web content, and conversational transcripts, makes it an excellent choice for training robust machine learning models capable of handling a wide range of contexts. The English portion of the NER dataset is divided into a 59,924/8,528/8,262 train/dev/test split.

\item BioNLP2004 \cite{collier-kim-2004-introduction}. Designed specifically for the bioinformatics and biomedical domains, the BioNLP2004 NER dataset was launched as part of the BioNLP Shared Task 2004 to evaluate systems that automatically identify named entities within biomedical texts. Comprising abstracts from the PubMed database, this dataset is meticulously annotated with gene and protein names, serving as a critical tool for developing and testing NER models in biological literature. It offers a 16,619/1,927/3,856 train/dev/test distribution, facilitating the training and evaluation phases of NER models in this specialized field.

\end{itemize}

\end{itemize}

\end{CJK*}

\end{document}